%% file: main.tex
\documentclass[10pt,twocolumn,letterpaper]{article}

\usepackage{iccv}
\usepackage{times}
\usepackage{epsfig}
\usepackage{graphicx}
\usepackage{amsmath}
\usepackage{amssymb}

\usepackage{multirow}
\usepackage{xcolor}
\usepackage{caption} 
\usepackage{pifont}

\usepackage{subcaption}
\usepackage{textcomp}
\usepackage{enumitem}
\usepackage{siunitx}
\usepackage{graphicx}

\usepackage{tabularx}

\makeatletter
\@namedef{ver@everyshi.sty}{}
\makeatother
\usepackage{pgfplots}

\usepackage{stfloats}

\input{definitions}

\usepackage[breaklinks=true,bookmarks=false]{hyperref}
\usepackage{url}
\hypersetup{
    colorlinks = true,
    urlcolor = blue,
    citecolor=.,
    linkcolor=.,
}
\usepackage{cleveref}

\iccvfinalcopy 

\def\iccvPaperID{1835} 

\begin{document}

\title{FIERY: Future Instance Prediction in Bird's-Eye View \\from Surround Monocular Cameras}

{
\centering
\author{Anthony Hu\textsuperscript{1,2}
\qquad Zak Murez\textsuperscript{1}
\qquad Nikhil Mohan\textsuperscript{1}
\qquad Sofía Dudas\textsuperscript{1}\\
\qquad Jeffrey Hawke\textsuperscript{1}
\qquad Vijay Badrinarayanan\textsuperscript{1}
\qquad Roberto Cipolla\textsuperscript{2}
\qquad Alex Kendall\textsuperscript{1}\\
\\
\textsuperscript{1}Wayve, UK. \qquad \textsuperscript{2}University of Cambridge, UK.


}}

\maketitle

\pagestyle{plain}
\ificcvfinal\thispagestyle{empty}\fi

\begin{abstract}
    Driving requires interacting with road agents and predicting their future behaviour in order to navigate safely. We present FIERY: a probabilistic future prediction model in bird's-eye view from monocular cameras. Our model predicts future instance segmentation and motion of dynamic agents that can be transformed into non-parametric future trajectories.
    Our approach combines the perception, sensor fusion and prediction components of a traditional autonomous driving stack by estimating bird's-eye-view prediction directly from surround RGB monocular camera inputs. FIERY learns to model the inherent stochastic nature of the future solely from camera driving data in an end-to-end manner, without relying on HD maps, and predicts multimodal future trajectories.
    We show that our model outperforms previous prediction baselines on the NuScenes and Lyft datasets. The code and trained models are available at \href{https://github.com/wayveai/fiery}{\texttt{https://github.com/wayveai/fiery}}.
\end{abstract}


\input{sections/1-introduction}

\input{sections/2-related_work}

\input{sections/3-model_description}

\input{sections/4-experiments}

\input{sections/5-results}

\input{sections/6-conclusion}

\clearpage
\input{sections/7-appendix.tex}

\clearpage
{\small
\bibliographystyle{ieee_fullname}
\bibliography{egbib}
}

\end{document}

%% file: definitions.tex
\makeatletter
\DeclareRobustCommand\onedot{\futurelet\@let@token\@onedot}
\def\@onedot{\ifx\@let@token.\else.\null\fi\xspace}

\def\ie{\emph{i.e}\onedot} 

\makeatletter
\def\thickhline{%
  \noalign{\ifnum0=`}\fi\hrule \@height \thickarrayrulewidth \futurelet
   \reserved@a\@xthickhline}
\def\@xthickhline{\ifx\reserved@a\thickhline
               \vskip\doublerulesep
               \vskip-\thickarrayrulewidth
             \fi
      \ifnum0=`{\fi}}
\makeatother

\newlength{\thickarrayrulewidth}
\makeatletter
\def\mediumhline{%
  \noalign{\ifnum0=`}\fi\hrule \@height \mediumarrayrulewidth \futurelet
   \reserved@a\@xthickhline}
\def\@xmediumhline{\ifx\reserved@a\mediumhline
               \vskip\doublerulesep
               \vskip-\mediumarrayrulewidth
             \fi
      \ifnum0=`{\fi}}
\makeatother

\newlength{\mediumarrayrulewidth}
\definecolor{blue1}{RGB}{0,176,227}
\definecolor{blue2}{RGB}{0,113,171}
\definecolor{blue3}{RGB}{40,80,140}

\definecolor{violet1}{RGB}{218,158,224}
\definecolor{violet2}{RGB}{137,72,155}
\definecolor{violet3}{RGB}{60,28,83}

\newcolumntype{Y}{>{\centering\arraybackslash}X}

%% file: sections/1-introduction.tex
\section{Introduction}

\begin{figure*}
\centering
    \begin{subfigure}[b]{0.62\textwidth}
    \centering
    \includegraphics[width=\linewidth]{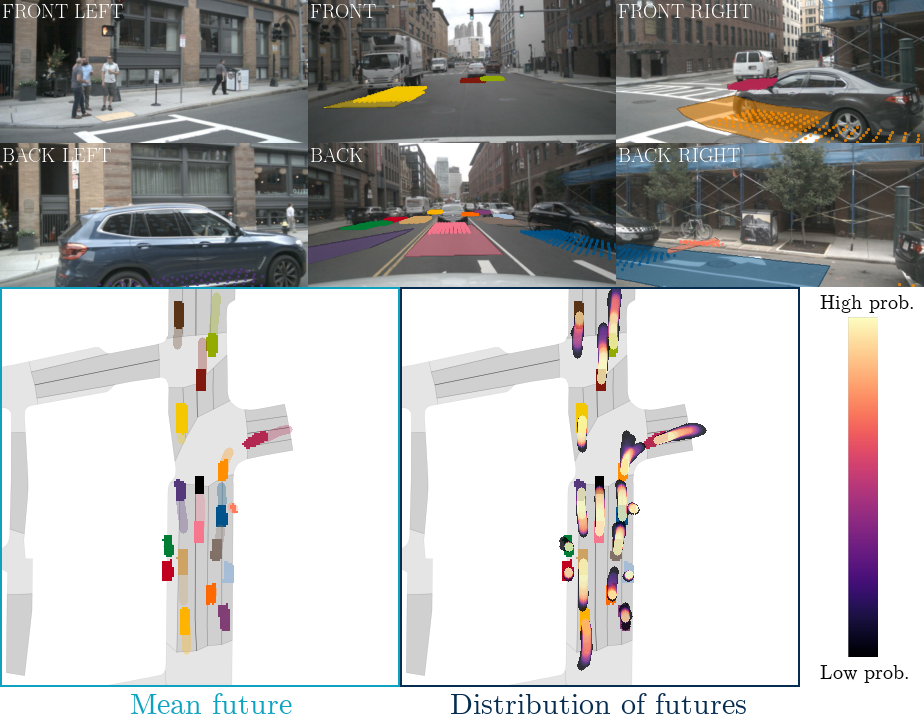}%
    \end{subfigure}
    \par\smallskip
    \begin{subfigure}[t]{0.62\textwidth}
    \includegraphics[width=\linewidth]{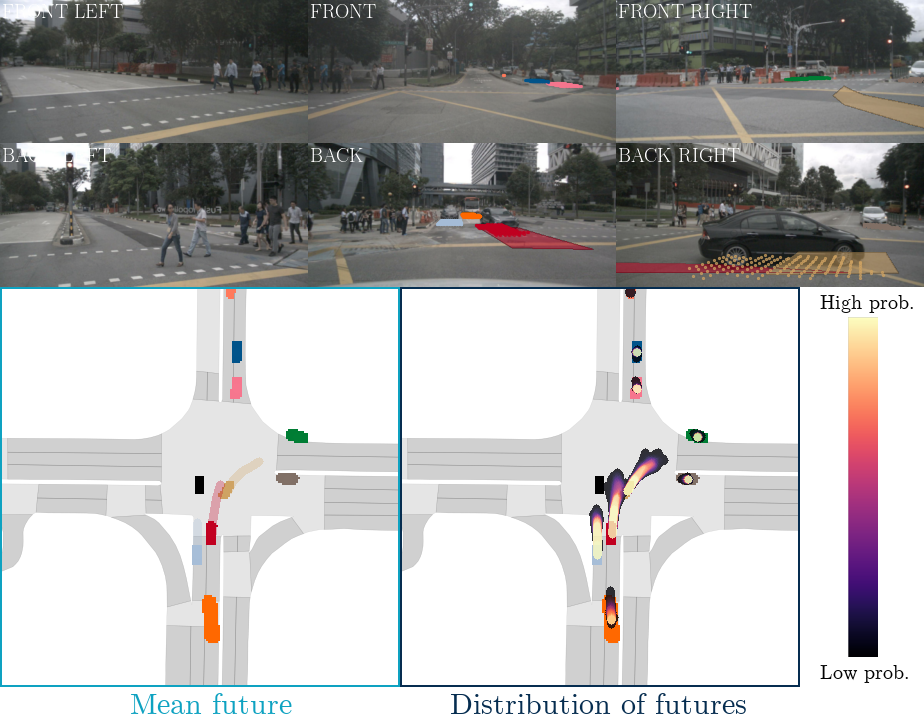}%
    \end{subfigure}
\caption{Multimodal future predictions by our bird's-eye view network.
Top two rows: RGB camera inputs. The predicted instance segmentations are projected to the ground plane in the images. We also visualise the mean future trajectory of dynamic agents as transparent paths.
Bottom row: future instance prediction in bird's-eye view in a $100\mathrm{m}\times100\mathrm{m}$ capture size around the ego-vehicle, which is indicated by a black rectangle in the center.}
\label{fig:teaser}
\end{figure*}

Prediction of future states is a key challenge in many autonomous decision making systems. This is particularly true for motion planning in highly dynamic environments: for example in autonomous driving where the motion of other road users and pedestrians has a substantial influence on the success of motion planning \cite{cui2019multimodal}. Estimating the motion and future poses of these road users enables motion planning algorithms to better resolve multimodal outcomes where the optimal action may be ambiguous knowing only the current state of the world.

Autonomous driving is inherently a geometric problem, where the goal is to navigate a vehicle safely and correctly through 3D space. As such, an orthographic bird's-eye view (BEV) perspective is commonly used for motion planning and prediction based on LiDAR sensing \cite{precog19, wu2020motionnet}. Recent advances in camera-based perception have rivalled LiDAR-based perception \cite{wang2019pseudo}, and we anticipate that this will also be possible for wider monocular vision tasks, including prediction. Building a perception and prediction system based on cameras would enable a leaner, cheaper and higher resolution visual recognition system over LiDAR sensing.

Most of the work in camera-based prediction to date has either been performed directly in the perspective view coordinate frame \cite{alahi2016social, jin2017predicting}, or using simplified BEV raster representations of the scene \cite{desire17, djuric2020uncertainty, cui2019multimodal} generated by HD-mapping systems such as \cite{liang2018deep, gao2020vectornet}. We wish to build predictive models that operate in an orthographic bird's-eye view frame (due to the benefits for planning and control \cite{7490340}), though \emph{without} relying on auxiliary systems to generate a BEV raster representation of the scene.

A key theme in robust perception systems for autonomous vehicles has been the concept of early sensor fusion, generating 3D object detections directly from image and LiDAR data rather than seeking to merge the predicted outputs of independent object detectors on each sensor input. Learning a task jointly from multiple sources of sensory data as in \cite{xu2018pointfusion}, rather than a staged pipeline, has been demonstrated to offer improvement to perception performance in tasks such as object detection. We seek similar benefits in joining perception and sensor fusion to prediction by estimating bird's-eye-view prediction directly from surround RGB monocular camera inputs, rather than a multi-stage discrete pipeline of tasks.

Lastly, traditional autonomous driving stacks \cite{ferguson08} tackle future prediction by extrapolating the current behaviour of dynamic agents, without taking into account possible interactions. They rely on HD maps and use road connectivity to generate a set of future trajectories. Instead, FIERY learns to predict future motion of road agents directly from camera driving data in an end-to-end manner, without relying on HD maps. It can reason about the probabilistic nature of the future, and predicts multimodal future trajectories (see \href{https://wayve.ai/blog/fiery-future-instance-prediction-birds-eye-view}{blog post} and \Cref{fig:teaser}).


To summarise the main contributions of this paper:
\begin{enumerate}[itemsep=1mm, parsep=0pt]
    \item We present the first future prediction model in bird's-eye view from monocular camera videos. Our framework explicitly reasons about multi-agent dynamics by predicting future instance segmentation and motion in bird's-eye view.
    \item Our probabilistic model predicts plausible and multi-modal futures of the dynamic environment.
    \item We demonstrate quantitative benchmarks for future dynamic scene segmentation, and show that our learned prediction outperforms previous prediction baselines for autonomous driving on the NuScenes \cite{nuscenes19} and Lyft \cite{lyft2019} datasets.
    
\end{enumerate}

%% file: sections/2-related_work.tex
\section{Related Work}
\paragraph{Bird's-eye view representation from cameras.}
Many prior works \cite{zhu2019generative, wang2019monocular} have tackled the inherently ill-posed problem \cite{groenendijk20} of lifting 2D perspective images into a bird's-eye view representation. \cite{Pan_2020,ng2020bevseg} dealt specifically with the problem of generating semantic BEV maps directly from images and used a simulator to obtain the ground truth.

Recent multi-sensor datasets, such as NuScenes \cite{nuscenes19} or Lyft \cite{lyft2019}, made it possible to directly supervise models on real-world data by generating bird's-eye view semantic segmentation labels from 3D object detections. \cite{roddick20} proposed a Bayesian occupancy network to predict road elements and dynamic agents in BEV directly from monocular RGB images.  Most similar to our approach, Lift-Splat \cite{philion20} learned a depth distribution over pixels to lift camera images to a 3D point cloud, and project the latter into BEV using camera geometry. Fishing Net \cite{hendy20} tackled the problem of predicting deterministic future bird's-eye view semantic segmentation using camera, radar and LiDAR inputs.

\paragraph{Future prediction.} Classical methods for future prediction generally employ a multi-stage detect-track-predict paradigm for trajectory prediction \cite{chai2019multipath, hong2019rules, tang2019multiple}.  However, these methods are prone to cascading errors and high latency, and thus many have turned to an end-to-end approach for future prediction.  Most end-to-end approaches rely heavily on LiDAR data \cite{Luo_2018_CVPR, djuric2020multixnet}, showing improvements by incorporating HD maps \cite{pmlr-v87-casas18a},  encoding constraints \cite{casas2019spatiallyaware}, and fusing radar and other sensors for robustness \cite{shah2020liranet}.  These end-to-end methods are faster and have higher performance as compared to the traditional multi-stage approaches.

The above methods attempt future prediction by producing a single deterministic trajectory \cite{pmlr-v87-casas18a, hendy20}, or a single distribution to model the uncertainty of each waypoint of the trajectory \cite{casas2019spatiallyaware, djuric2020multixnet}.  However, in the case of autonomous driving, one must be able to anticipate a range of behaviors for actors in the scene, jointly.  From an observed past, there are many valid and probable futures that could occur \cite{hu2020probabilistic}.  Other work \cite{chai2019multipath, tang2019multiple, phanminh2020covernet} has been done on probabilistic multi-hypothesis trajectory prediction, however all assume access to top-down rasterised representations as inputs. Our approach is the first to predict diverse and plausible future vehicle trajectories directly from raw camera video inputs.

%% file: sections/3-model_description.tex
\begin{figure*}[h]
    \captionsetup{singlelinecheck=off}
    \centering
    \includegraphics[width=\linewidth]{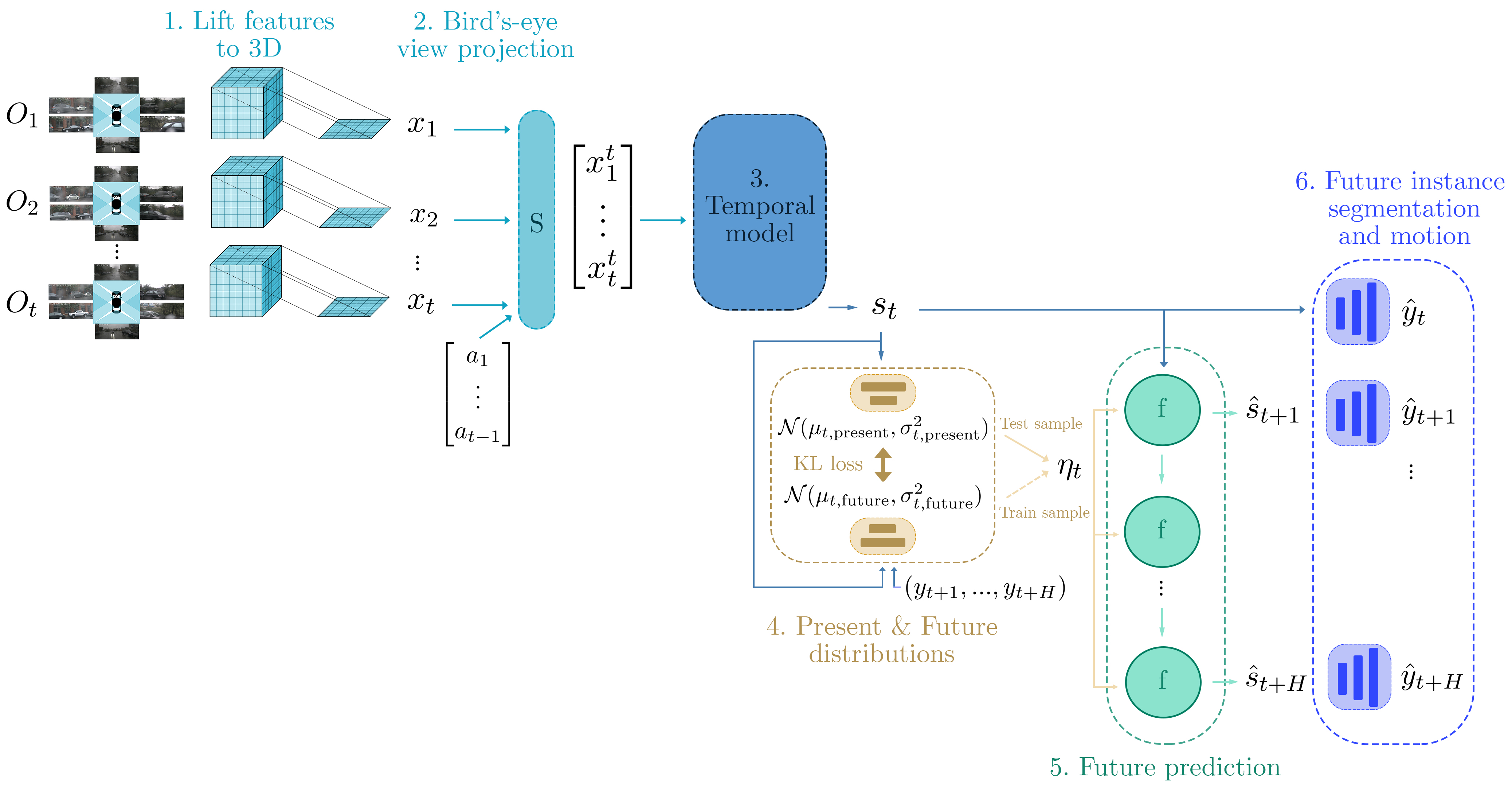}%
    \caption[]{The architecture of FIERY: a future prediction model in bird's-eye view from camera inputs.
    \begin{enumerate}[itemsep=0.3mm, parsep=0pt]
        \item At each past timestep $\{1,...,t\}$, we lift camera inputs $(O_1, ..., O_t)$ to 3D by predicting a depth probability distribution over pixels and using known camera intrinsics and extrinsics.
        \item These features are projected to bird's-eye view $(x_1, ..., x_t)$. Using past ego-motion $(a_1, ..., a_{t-1})$, we transform the bird's-eye view features into the present reference frame (time $t$) with a Spatial Transformer module $S$.
        \item A 3D convolutional temporal model learns a spatio-temporal state $s_t$.
        \item We parametrise two probability distributions: the present and the future distribution. The present distribution is conditioned on the current state $s_t$, and the future distribution is conditioned on both the current state $s_t$ and future labels $(y_{t+1}, ..., y_{t+H})$.
        \item We sample a latent code $\eta_t$ from the future distribution during training, and from the present distribution during inference. The current state $s_t$ and the latent code $\eta_t$ are the inputs to the future prediction model that recursively predicts future states ($\hat{s}_{t+1}, ..., \hat{s}_{t+H})$.
        \item The states are decoded into future instance segmentation and future motion in bird's-eye view $(\hat{y}_t, ..., \hat{y}_{t+H})$.
    \end{enumerate}
    }
    \vspace{-14pt}
    \label{fig:model-diagram}
\end{figure*}

\section{Model Architecture}
An overview of our model is given in \Cref{fig:model-diagram}.

\subsection{Lifting camera features to 3D}
For every past timestep, we use the method of \cite{philion20} to extract image features from each camera and then lift and fuse them into a BEV feature map. In particular, each image is passed through a standard convolutional encoder $E$ (we use EfficientNet \cite{tan19} in our implementation) to obtain a set of features to be lifted and a set of discrete depth probabilities. Let $O_t = \{I_t^1, ..., I_t^n\}$ be the set of $n=6$ camera images at time $t$. We encode each image $I_t^k$ with the encoder: $e_t^k = E(I_t^k) \in \mathbb{R}^{(C+D)\times H_e \times W_e}$, with $C$ the number of feature channels, $D$ the number of discrete depth values and $(H_e, W_e)$ the feature spatial size. $D$ is equal to the number of equally spaced depth slices between $D_{\text{min}}$ (the minimum depth value) and $D_{\text{max}}$ (the maximum depth value) with size $D_{\mathrm{size}}=1.0\mathrm{m}$. Let us split this feature into two: $e_t^k = (e_{t,C}^k, e_{t,D}^k)$ with $e_{t,C}^k \in \mathbb{R}^{C\times H_e \times W_e}$ and $e_{t,D}^k \in \mathbb{R}^{D\times H_e \times W_e}$.  A tensor $u_t^k \in \mathbb{R}^{C \times D \times H_e \times W_e}$ is formed by taking the outer product of the features to be lifted with the depth probabilities:
\begin{equation}
    u_t^k = e_{t,C}^k \otimes e_{t,D}^k
\end{equation}

The depth probabilities act as a form of self-attention, modulating the features according to which depth plane they are predicted to belong to.
Using the known camera intrinsics and extrinsics (position of the cameras with respect to the center of gravity of the vehicle), these tensors from each camera $(u_t^1, ..., u_t^n)$ are lifted to 3D in a common reference frame (the inertial center of the ego-vehicle at time $t$). 

\subsection{Projecting to bird's-eye view}
In our experiments, to obtain a bird's-eye view feature, we discretise the space in $0.50\mathrm{m}\times 0.50\mathrm{m}$ columns in a $100\mathrm{m}\times100\mathrm{m}$ capture size around the ego-vehicle. The 3D features are sum pooled along the vertical dimension to form bird's-eye view feature maps $x_t \in \mathbb{R}^{C\times H \times W}$, with $(H, W) = (200, 200)$ the spatial extent of the BEV feature.

\subsection{Learning a temporal representation}
The past bird's-eye view features $(x_1, ..., x_t)$ are transformed to the present's reference frame (time $t$) using known past ego-motion $(a_1,...,a_{t-1})$. $a_{t-1} \in SE(3)$ corresponds to the ego-motion from $t-1$ to $t$, i.e. the translation and rotation of the ego-vehicle. Using a Spatial Transformer \cite{jaderberg15} module $S$, we warp past features $x_i$ to the present for $i \in \{1,...,t-1\}$:

\begin{equation}
    x_i^t = S(x_i, a_{t-1}\cdot a_{t-2} \cdot ... \cdot a_i)
\end{equation}

Since we lose the past ego-motion information with this operation, we concatenate spatially-broadcast actions to the warped past features $x_i^t$.

These features are then the input to a temporal model $\mathcal{T}$ which outputs a spatio-temporal state $s_t$:
\begin{equation}
    s_t = \mathcal{T}(x_1^t, x_2^t, ..., x_t^t)
\end{equation}

with $x_t^t=x_t$. $\mathcal{T}$ is a 3D convolutional network with local spatio-temporal convolutions, global 3D pooling layers, and skip connections. For more details on the temporal module, see \Cref{appendix:model}.

\subsection{Present and future distributions}
Following \cite{hu2020probabilistic} we adopt a conditional variational approach to model the inherent stochasticity of future prediction. We introduce two distributions: a \emph{present distribution} $P$ which only has access to the current spatio-temporal state $s_t$, and a \emph{future distribution} $F$ that additionally has access to the observed future labels $(y_{t+1},..., y_{t+H})$, with $H$ the future prediction horizon. The labels correspond to future centerness, offset, segmentation, and flow (see \Cref{section-temporally-consistent}).

We parametrise both distributions as diagonal Gaussians with mean $\mu \in \mathbb{R}^L$ and variance $\sigma^2 \in \mathbb{R}^L$, $L$ being the latent dimension. During training, we use samples $\eta_t \sim \mathcal{N}(\mu_{t,\text{future}}, \sigma_{t,\text{future}}^2)$ from the future distribution to enforce predictions consistent with the observed future, and a mode covering Kullback-Leibler divergence loss to encourage the present distribution to cover the observed futures:

\begin{equation}
    L_{\text{probabilistic}} = D_\text{KL}(F(\cdot| s_t, y_{t+1}, ..., y_{t+H}) ~ || ~P(\cdot | s_t))
    \label{eq:kl-loss}
\end{equation}

During inference, we sample $\eta_t \sim \mathcal{N}(\mu_{t,\text{present}}, \sigma_{t,\text{present}}^2)$ from the present distribution where each sample encodes a possible future.

\subsection{Future prediction in bird's-eye view}

The future prediction model is a convolutional gated recurrent unit network taking as input the current state $s_t$ and the latent code $\eta_t$ sampled from the future distribution $F$ during training, or the present distribution $P$ for inference. It recursively predicts future states $(\hat{s}_{t+1}, ..., \hat{s}_{t+H})$.

\subsection{Future instance segmentation and motion}
\label{section-temporally-consistent}
The resulting features are the inputs to a bird's-eye view decoder $\mathcal{D}$ which has multiple output heads: semantic segmentation, instance centerness and instance offset (similarily to \cite{cheng20}), and future instance flow. For $j \in \{0, ..., H\}$:

\begin{equation}
    \hat{y}_{t+j} = \mathcal{D}(\hat{s}_{t+j})
\end{equation}

with $\hat{s}_t = s_t$.

\begin{figure*}
\centering
    \begin{subfigure}[b]{\textwidth}
    \centering
    \includegraphics[width=0.6\linewidth]{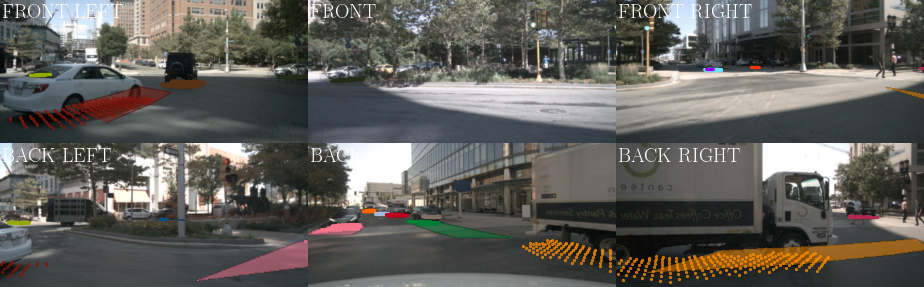}%
    \caption{Camera inputs.}
    \label{subfig-camera}
    \end{subfigure}
    \begin{subfigure}[b]{0.19\textwidth}
    \includegraphics[width=\linewidth]{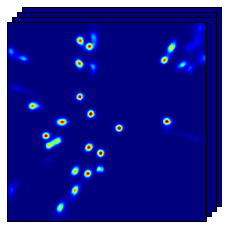}%
    \caption{Centerness.}
    \label{subfig-centerness}
    \end{subfigure}
    \begin{subfigure}[b]{0.19\textwidth}
    \includegraphics[width=\linewidth]{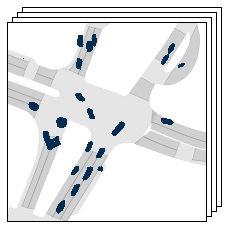}%
    \caption{Segmentation.}
    \label{subfig-segmentation}
    \end{subfigure}
    \begin{subfigure}[b]{0.19\textwidth}
    \includegraphics[width=\linewidth]{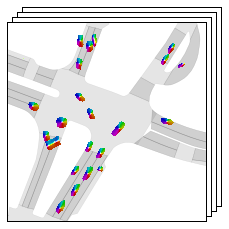}%
    \caption{Offset.}
    \label{subfig-offset}
    \end{subfigure}
    \begin{subfigure}[b]{0.19\textwidth}
    \includegraphics[width=\linewidth]{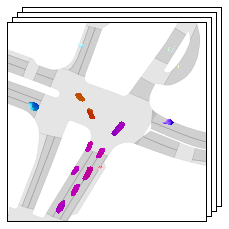}%
    \caption{Future flow.}
    \label{subfig-flow}
    \end{subfigure}
    \begin{subfigure}[b]{0.19\textwidth}
    \includegraphics[width=\linewidth]{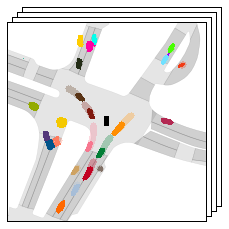}%
    \caption{Instance segmentation.}
    \label{subfig-instance}
    \end{subfigure}
\caption{Outputs from our model. (b) shows a heatmap of instance centerness and indicates the probability of finding an instance center (from blue to red). (c) represents the vehicles segmentation. (d) shows a vector field indicating the direction to the instance center. (e) corresponds to future motion  -- notice how consistent the flow is for a given instance, since it's a rigid-body motion. (f) shows the final output of our model: a sequence of temporally consistent future instance segmentation in bird's-eye view where: (i) Instance centers are obtained by non-maximum suppression. (ii) The pixels are then grouped to their closest instance center using the offset vector. (iii) Future flow allows for consistent instance identification by comparing the warped centers using future flow from $t$ to $t+1$, and the centers at time $t+1$. The ego-vehicle is indicated by a black rectangle.}
\label{fig:model-outputs}
\end{figure*}

For each future timestep $t+j$, the instance centerness indicates the probability of finding an instance center (see \Cref{subfig-centerness}). By running non-maximum suppression, we get a set of instance centers. The offset is a vector pointing to the center of the instance (\Cref{subfig-offset}), and can be used jointly with the segmentation map (\Cref{subfig-segmentation}) to assign neighbouring pixels to its nearest instance center and form the bird's-eye view instance segmentation (\Cref{subfig-instance}). The future flow (\Cref{subfig-flow}) is a displacement vector field of the dynamic agents. It is used to consistently track instances over time by comparing the flow-warped instance centers at time $t+j$ and the detected instance centers at time $t+j+1$ and running a Hungarian matching algorithm \cite{hungarian55}.

A full description of our model is given in \Cref{appendix:model}.

\subsection{Losses}
For semantic segmentation, we use a top-$k$ cross-entropy loss \cite{wang2019pseudo}. As the bird's-eye view image is largely dominated by the background, we only backpropagate the top-$k$ hardest pixels. In our experiments, we set $k=25\%$. The centerness loss is a $\ell_2$ distance, and both offset and flow losses are $\ell_1$ distances. We exponentially discount future timesteps with a parameter $\gamma=0.95$.

%% file: sections/4-experiments.tex
\section{Experimental Setting}
\subsection{Dataset}
We evaluate our approach on the NuScenes \cite{nuscenes19} and Lyft \cite{lyft2019} datasets. NuScenes contains $1000$ scenes, each $20$ seconds in length, annotated at $2$Hz. The Lyft dataset contains $180$ scenes, each $25-45$ seconds in length, annotated at $5$Hz. In both datasets, the camera rig covers the full $360$\textdegree~field of view around the ego-vehicle, and is comprised of $6$ cameras with a small overlap in field of view. Camera intrinsics and extrinsics are available for each camera in every scene.

The labels $(y_t, ..., y_{t+H})$ are generated by projecting the provided 3D bounding boxes of vehicles into the bird's-eye view plane to create a bird's-eye view occupancy grid. See \Cref{appendix:dataset} for more details. All the labels $(y_t, ..., y_{t+H})$ are in the present's reference frame and are obtained by transforming the labels with the ground truth future ego-motion.

\subsection{Metrics}
\paragraph{Future Video Panoptic Quality.} We want to measure the performance of our system in both:
\begin{enumerate}[label=(\roman*), itemsep=0.5mm, parsep=0pt]
    \item Recognition quality: how consistently the instances are detected over time.
    \item Segmentation quality: how accurate the instance segmentations are.
\end{enumerate}

We use the \emph{Video Panoptic Quality} (VQP) \cite{kim2020vps} metric defined as:

\begin{equation}
    \text{VPQ} = \sum_{t=0}^H \frac{\sum_{(p_t,q_t) \in TP_t} \text{IoU}(p_t,q_t)}{|TP_t| + \frac{1}{2}|FP_t| + \frac{1}{2}|FN_t|}
\end{equation}

with $TP_t$ the set of true positives at timestep $t$ (correctly detected ground truth instances), $FP_t$ the set of false positives at timestep $t$ (predicted instances that do not match any ground truth instance), and $FN_t$ the set of false negatives at timestep $t$ (ground truth instances that were not detected). A true positive corresponds to a predicted instance segmentation that has: (i) an intersection-over-union (IoU) over 0.5 with the ground truth, and (ii) an instance id that is consistent with the ground truth over time (correctly tracked).







\paragraph{Generalised Energy Distance.} 
To measure the ability of our model to predict multi-modal futures, we report the \emph{Generalised Energy Distance} ($D_{\text{GED}}$) \cite{szekely17} (defined in \Cref{appendix:probabilistic}).

\subsection{Training}
Our model takes $1.0$s of past context and predicts $2.0$s in the future. In NuScenes, this corresponds to $3$ frames of past temporal context and $4$ frames into the future at 2Hz. In the Lyft dataset, this corresponds to $6$ frames of past context and $10$ frames in the future at 5Hz.

For each past timestep, our model processes $6$ camera images at resolution $224\times480$. It outputs a sequence of $100\mathrm{m}\times100\mathrm{m}$ BEV predictions at $50\mathrm{cm}$ pixel resolution in both the $x$ and $y$ directions resulting in a bird's-eye view video with spatial dimension $200\times200$. We use the Adam optimiser with a constant learning rate of $3\times 10^{-4}$. We train our model on $4$ Tesla V100 GPUs with a batch size of $12$ for $20$ epochs at mixed precision.

%% file: sections/5-results.tex
\section{Results}
\subsection{Comparison to the literature}
Since predicting future instance segmentation in bird's-eye view is a new task, we begin by comparing our model to previous published methods on bird's-eye view semantic segmentation from monocular cameras.

Many previous works \cite{lu19-icra-ral,Pan_2020,roddick20,philion20,saha21} have proposed a model to output the dynamic scene bird's-eye view segmentation from multiview camera images of a single timeframe. For comparison, we adapt our model so that the past context is reduced to a single observation, and we set the future horizon $H=0$ (to only predict the present's segmentation). We call this model \emph{FIERY Static} and report the results in \Cref{table:single-timestep}. We observe that FIERY Static outperforms all previous baselines.

\begin{table}[h!]
\centering
\begin{tabularx}{0.48\textwidth}{l|YYY}
\thickhline
\multicolumn{1}{c|}{} & \multicolumn{3}{c}{\color{darkgray}{\textbf{Intersection-over-Union (IoU)}}}\\
& Setting 1 & Setting 2 & Setting 3\\
\mediumhline
VED \cite{lu19-icra-ral} & 8.8 & - & -\\
PON \cite{roddick20} & 24.7 & - & -\\
VPN \cite{Pan_2020} & 25.5 & - & -\\
STA \cite{saha21} & 36.0 & - & -\\
Lift-Splat \cite{philion20} & - & 32.1 & - \\
Fishing Camera \cite{hendy20} & - & - & 30.0\\
Fishing Lidar \cite{hendy20} & - & - & 44.3\\
FIERY Static & 37.7& 35.8& - \\
\textbf{FIERY} & \textbf{39.9}& \textbf{38.2}& \textbf{57.6}\\
\thickhline
\end{tabularx}
\caption{Bird's-eye view semantic segmentation on NuScenes in the settings of the respective published methods. \\Setting 1: $100\mathrm{m}\times50\mathrm{m}$ at $25$cm resolution. Prediction of the present timeframe. \\Setting 2: $100\mathrm{m}\times100\mathrm{m}$ at $50$cm resolution. Prediction of the present timeframe. \\Setting 3: $32.0\mathrm{m}\times19.2\mathrm{m}$ at $10$cm resolution. Prediction $2.0$s in the future. In this last setting we compare our model to two variants of Fishing Net \cite{hendy20}: one using camera inputs, and one using LiDAR inputs.}
\label{table:single-timestep}
\end{table}

We also train a model that takes 1.0s of past observations as context (\emph{FIERY}) and note that it achieves an even higher intersection-over-union over its single-timeframe counterpart that has no past context. This is due to our model's ability to accumulate information over time and better handle partial observability and occlusions (see \Cref{fig:temporal-fusion}). Qualitatively, as shown in \Cref{fig:qual-literature}, our predictions are much sharper and more accurate.

Finally, we compare our model to Fishing Net \cite{hendy20}, where the authors predicts bird's-eye view semantic segmentation $2.0\mathrm{s}$ in the future. Fishing Net proposes two variants of their model: one using camera as inputs, and one using LiDAR as inputs. FIERY performs much better than both the camera and LiDAR models, hinting that computer vision networks are starting to become competitive with LiDAR sensing for the prediction task.

\subsection{Future instance prediction}
In order to compare the performance of our model for future instance segmentation and motion prediction, we introduce the following baselines:

\paragraph{Static model.} The most simple approach to model dynamic obstacles is to assume that they will not move and remain static. We use FIERY Static to predict the instance segmentation of the present timestep (time $t$), and repeat this prediction in the future. We call this baseline the \emph{Static model} as it should correctly detect all static vehicles, since the future labels are in the present's reference frame.
\paragraph{Extrapolation model.} Classical prediction methods \cite{fiorini98,fraichard03} extrapolate the current behaviour of dynamic agents in the future. We run FIERY Static on every past timesteps to obtain a sequence of past instance segmentations. We re-identify past instances by comparing the instance centers and running a Hungarian matching algorithm. We then obtain past trajectories of detected vehicles, which we extrapolate in the future and transform the present segmentation accordingly.
\paragraph{}
We also report the results of various ablations of our proposed architecture:
\begin{itemize}[itemsep=0.5mm, parsep=0pt]
\item \textbf{No temporal context.} This model only uses the features $x_t$ from the present timestep to predict the future (\ie we set the 3D convolutional temporal model to the identity function).

\begin{table*}[th!]
\centering
\begin{tabularx}{0.65\textwidth}{l|YY|YY}
\thickhline
    \multicolumn{1}{c|}{} 
& \multicolumn{2}{c|}{\color{darkgray}{\textbf{Intersection-over-Union}}}
& \multicolumn{2}{c}{\color{darkgray}{\textbf{Video Panoptic Quality}}}\\ 

& Short & Long & Short & Long \\
\mediumhline
Static model & 47.9 & 30.3 & 43.1 & 24.5 \\
Extrapolation model  & 49.2 & 30.8  & 43.8 & 24.9 \\
\hline
No temporal context & 51.7 & 32.6 & 40.3 & 24.1 \\
No transformation  & 53.0 & 33.8 & 41.7 & 24.6\\
No unrolling & 55.4 & 34.9 & 44.2 & 26.2 \\
No future flow  & 58.0 & 36.7  & 44.6 & 26.9\\
Uniform depth  & 57.1 & 36.2 & 46.8 & 27.8\\
Deterministic & 58.2 & 36.6  & 48.3 & 28.5\\
\textbf{FIERY}  & \textbf{59.4} & \textbf{36.7} & \textbf{50.2} & \textbf{29.9} \\
\thickhline
\end{tabularx}
\caption{Future instance segmentation in bird's-eye view for $2.0\mathrm{s}$ in the future on NuScenes. We report future Intersection-over-Union (IoU) and Video Panoptic Quality (VPQ), evaluated at different ranges: $30\mathrm{m}\times30\mathrm{m}$ (Short) and $100\mathrm{m}\times100\mathrm{m}$ (Long) around the ego-vehicle. Results are reported as percentages.}
\label{table:nuscenes}
\vspace{-10pt}
\end{table*}

\item \textbf{No transformation.} Past bird's-eye view features $(x_1, ..., x_t)$ are not warped to the present's reference frame.

\item \textbf{No future flow.} This model does not predict future flow.

\item \textbf{No unrolling.} Instead of recursively predicting the next states $\hat{s}_{t+j}$ and decoding the corresponding instance information $\hat{y}_{t+j} = \mathcal{D}(\hat{s}_{t+j})$, this variant directly predicts all future instance centerness, offset, segmentation and flow from $s_t$.

\item \textbf{Uniform depth.} We lift the features from the encoder $(e_t^1, ..., e_t^n)$ with the Orthographic Feature Transform \cite{roddick19} module. This corresponds to setting the depth probability distribution to a uniform distribution.

\item \textbf{Deterministic.} No probabilistic modelling.
\end{itemize}

We report the results in \Cref{table:nuscenes} (on NuScenes) and \Cref{table:lyft} (on Lyft) of the mean prediction of our probabilistic model (i.e. we set the latent code $\eta_t$ to the mean of the present distribution: $\eta_t = \mu_{t, \text{present}}$).

\subsection{Analysis}

\definecolor{grad1}{HTML}{AF2624}
\definecolor{grad2}{HTML}{912E3A}
\definecolor{grad3}{HTML}{753752}
\definecolor{grad4}{HTML}{5A406A}
\definecolor{grad5}{HTML}{434881}
\definecolor{grad6}{HTML}{2F4F97}
\definecolor{grad7}{HTML}{155AB6}
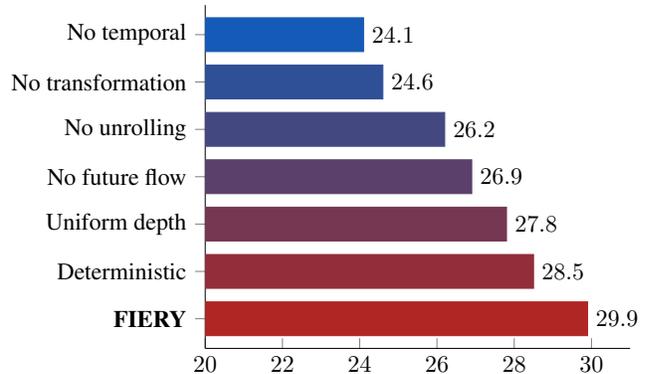
\begin{figure}[h]
\centering
\begin{tikzpicture}[scale=0.9]
\begin{axis}[
    xbar=0pt,
    /pgf/bar shift=0pt,
    legend style={
    legend columns=4,
        at={(xticklabel cs:0.5)},
        anchor=north,
        draw=none
    },
    ytick={0,...,6},
    axis y line*=left, 
    axis x line*=bottom, 
    xtick={20, 22,24,26,28,30},
    width=.45\textwidth,
    bar width=5mm,
    xmin=20,
    xmax=31,
    y=7mm, 
    enlarge y limits={abs=0.625},
    yticklabels={{\textbf{FIERY}}, 
    {Deterministic}, 
    {Uniform depth},
    {No future flow}, 
    {No unrolling}, 
    {No transformation},
    {No temporal}},
    nodes near coords,
    nodes near coords style={text=black},
    every axis plot/.append style={fill}
]
\addplot[grad1] coordinates {(29.9,0)};
\addplot[grad2] coordinates {(28.5,1)};
\addplot[grad3] coordinates {(27.8,2)};
\addplot[grad4] coordinates {(26.9,3)};
\addplot[grad5] coordinates {(26.2,4)};
\addplot[grad6] coordinates {(24.6,5)};
\addplot[grad7] coordinates {(24.1,6)};
\end{axis}  
\end{tikzpicture}
\caption{Performance comparison of various ablations of our model. We measure future Video Panoptic Quality $2.0\mathrm{s}$ in the future on NuScenes.}
\label{fig:barplot-results}
\end{figure}

\begin{table}[th!]
\centering
\begin{tabularx}{0.48\textwidth}{l|YY}
\thickhline
    \multicolumn{1}{c|}{} 
& \multicolumn{2}{c}{\color{darkgray}{\textbf{IoU$\mid$VPQ}}}\\ 

& Short & Long \\
\mediumhline
Static model &35.3$\mid$36.4& 24.1$\mid$20.7\\
Extrapolation model & 37.4$\mid$37.5 & 24.8$\mid$21.2\\
\textbf{FIERY} &  \textbf{57.8$\mid$50.0}& \textbf{36.3$\mid$29.2}\\
\thickhline
\end{tabularx}
\caption{Future instance prediction in bird's-eye view for $2.0$s in the future on the Lyft dataset. We report future Intersection-over-Union and Video Panoptic Quality.}
\label{table:lyft}
\end{table}

FIERY largely outperforms the Static and Extrapolation baselines for the task of future prediction. 
\Cref{fig:barplot-results} shows the performance boost our model gains from different parts of the model.

\paragraph{Temporal model.} The \emph{No temporal context} variant performs similarly to the static model. That is to be expected as this model does not have any information from the past, and cannot infer much about the motion of road agents.

\paragraph{Transformation to the present's reference frame.} There is a large performance drop when we do not transform past features to the present's reference frame. This can be explained by how much easier it is for the temporal model to learn correspondences between dynamic vehicles when the ego-motion is factored out.

Past prediction models either naively fed past images to a temporal model \cite{ballas16,hu2020probabilistic}, or did not use a temporal model altogether and simply concatenated past features \cite{luc17, hendy20}. We believe that in order to learn temporal correspondences, past features have to be mapped to a common reference frame and fed into a high capacity temporal model, such as our proposed 3D convolutional architecture.

\paragraph{Predicting future states.} When predicting the future, it is important to model its sequential nature, i.e. the prediction at time $t+j+1$ should be conditioned on the prediction at time $t+j$.

The \emph{No unrolling} variant which directly predicts all future instance segmentations and motions from the current state $s_t$, results in a large performance drop. This is because the sequential constraint is no longer enforced, contrarily to our approach that predicts future states in a recursive way.

\paragraph{Future motion.} Learning to predict future motion allows our model to re-identify instances using the predicted flow and comparing instance centers. Our model is the first to produce temporally consistent future instance segmentation in bird's-eye view of dynamic agents. Without future flow, the predictions are no longer temporally consistent explaining the sharp decrease in performance. 

\paragraph{Lifting the features to 3D} Using a perfect depth model we could directly lift each pixel to its correct location in 3D space. Since our depth prediction is uncertain, we instead lift the features at different possible depth locations and assign a probability mass at each location, similar to \cite{philion20}. The \emph{Uniform depth} baseline uses the Orthographic Feature Transform to lift features in 3D, by setting a uniform distribution on all depth positions. We observe that such a naive lifting performs worse compared to a learned weighting over depth.

\paragraph{Present and future distributions.} A deterministic model has a hard task at hand. It has to output with full confidence which future will happen, even though the said future is uncertain. In our probabilistic setting, the model is guided during training with the future distribution that outputs a latent code that indicates the correct future. It also encourages the present distribution to cover the modes of the future distribution. This paradigm allows FIERY to predict both accurate and diverse futures as we will see in section \Cref{section-probabilistic}.

\paragraph{}
Further analyses on understanding the structure of the learned latent space and on the temporal horizon of future prediction is available in \Cref{appendix-additional-results}.

\subsection{Probabilistic modelling}
\label{section-probabilistic}

We compare our probabilistic future prediction model to the following baselines: M-Head, Bayesian Dropout and Classical VAE (more details in \Cref{appendix:probabilistic}). We report the results in \Cref{table:diversity}, and observe that our model predicts the most accurate and diverse futures.

%% file: sections/6-conclusion.tex
\section{Conclusion}
Autonomous driving requires decision making in multimodal scenarios, where the present state of the world is not always sufficient to reason correctly alone. Predictive models estimating the future state of the world -- particularly other dynamic agents -- are therefore a key component to robust driving. We presented the first prediction model of dynamic agents for autonomous driving in bird's-eye view from surround RGB videos. We posed this as an end-to-end learning problem in which our network models future stochasticity with a variational distribution. We demonstrated that FIERY predicts temporally consistent future instance segmentations and motion and is able to model diverse futures accurately. In future work, we would like to jointly train a driving policy to condition the future prediction model on future actions.
\paragraph{}
{\bf Acknowledgements} We are grateful to Giulio D'Ippolito, Hannes Liik and Piotr Sokólski for technical assistance. We also thank Juba Nait Saada, Oumayma Bounou, Brady Zhou and the reviewers for their insightful comments and suggestions on the manuscript. This work was partially supported by Toshiba Europe, grant G100453.

%% file: sections/7-appendix.tex
\appendix
\section{Additional Results}
\label{appendix-additional-results}

\subsection{Comparison with published methods}
\begin{figure*}
\centering
    \begin{subfigure}[c]{0.25\textwidth}
    \centering
    \includegraphics[width=\linewidth]{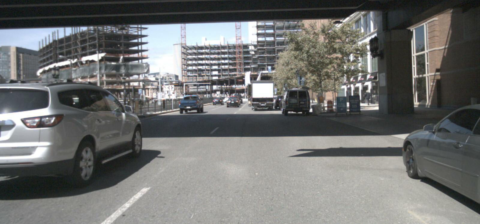}%
    \end{subfigure}
    \begin{subfigure}[c]{0.119\textwidth}
    \includegraphics[width=\linewidth]{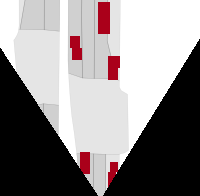}%
    \end{subfigure}
    \begin{subfigure}[c]{0.119\textwidth}
    \includegraphics[width=\linewidth]{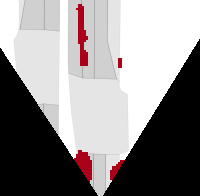}%
    \end{subfigure}
    \begin{subfigure}[c]{0.119\textwidth}
    \includegraphics[width=\linewidth]{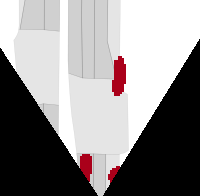}%
    \end{subfigure}
    \begin{subfigure}[c]{0.119\textwidth}
    \includegraphics[width=\linewidth]{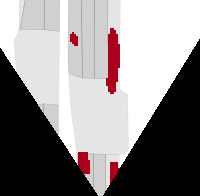}%
    \end{subfigure}
    \begin{subfigure}[c]{0.119\textwidth}
    \includegraphics[width=\linewidth]{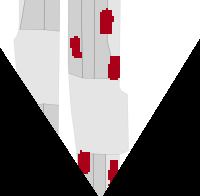}%
    \end{subfigure}
    \begin{subfigure}[c]{0.119\textwidth}
    \includegraphics[width=\linewidth]{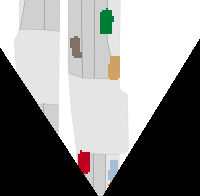}%
    \end{subfigure}
    \par\smallskip
    \begin{subfigure}[c]{0.25\textwidth}
    \centering
    \includegraphics[width=\linewidth]{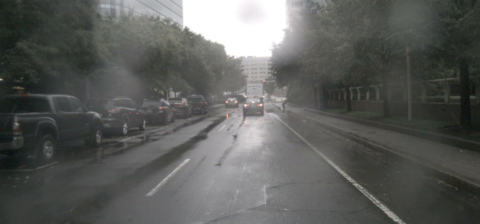}%
    \end{subfigure}
    \begin{subfigure}[c]{0.119\textwidth}
    \includegraphics[width=\linewidth]{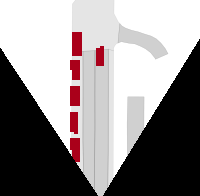}%
    \end{subfigure}
    \begin{subfigure}[c]{0.119\textwidth}
    \includegraphics[width=\linewidth]{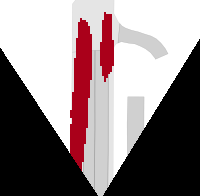}%
    \end{subfigure}
    \begin{subfigure}[c]{0.119\textwidth}
    \includegraphics[width=\linewidth]{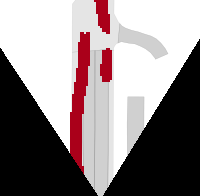}%
    \end{subfigure}
    \begin{subfigure}[c]{0.119\textwidth}
    \includegraphics[width=\linewidth]{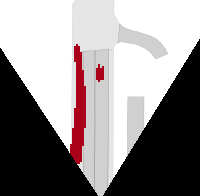}%
    \end{subfigure}
    \begin{subfigure}[c]{0.119\textwidth}
    \includegraphics[width=\linewidth]{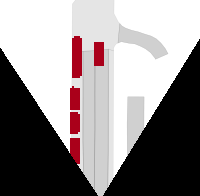}%
    \end{subfigure}
    \begin{subfigure}[c]{0.119\textwidth}
    \includegraphics[width=\linewidth]{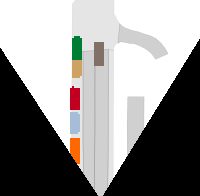}%
    \end{subfigure}

    \par\smallskip
    \begin{subfigure}[c]{0.25\textwidth}
    \centering
    \includegraphics[width=\linewidth]{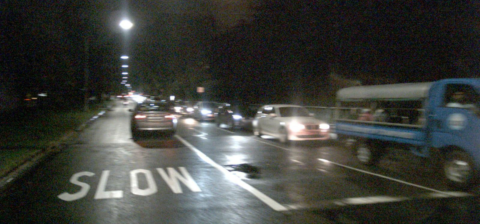}%
    \end{subfigure}
    \begin{subfigure}[c]{0.119\textwidth}
    \includegraphics[width=\linewidth]{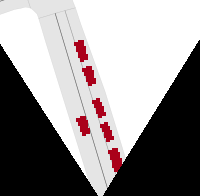}%
    \end{subfigure}
    \begin{subfigure}[c]{0.119\textwidth}
    \includegraphics[width=\linewidth]{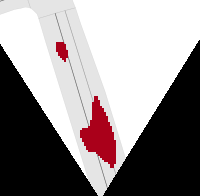}%
    \end{subfigure}
    \begin{subfigure}[c]{0.119\textwidth}
    \includegraphics[width=\linewidth]{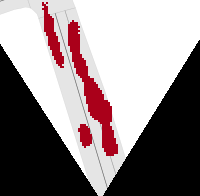}%
    \end{subfigure}
    \begin{subfigure}[c]{0.119\textwidth}
    \includegraphics[width=\linewidth]{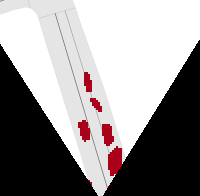}%
    \end{subfigure}
    \begin{subfigure}[c]{0.119\textwidth}
    \includegraphics[width=\linewidth]{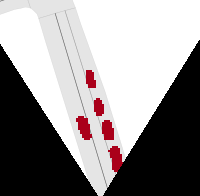}%
    \end{subfigure}
    \begin{subfigure}[c]{0.119\textwidth}
    \includegraphics[width=\linewidth]{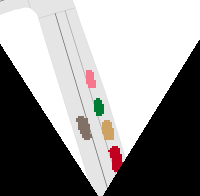}%
    \end{subfigure}

    \par\smallskip
    \begin{subfigure}[c]{0.25\textwidth}
    \centering
    \includegraphics[width=\linewidth]{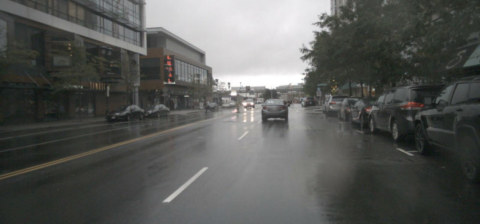}%
    \end{subfigure}
    \begin{subfigure}[c]{0.119\textwidth}
    \includegraphics[width=\linewidth]{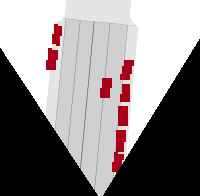}%
    \end{subfigure}
    \begin{subfigure}[c]{0.119\textwidth}
    \includegraphics[width=\linewidth]{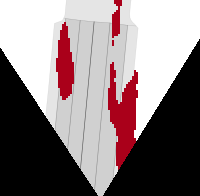}%
    \end{subfigure}
    \begin{subfigure}[c]{0.119\textwidth}
    \includegraphics[width=\linewidth]{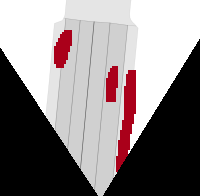}%
    \end{subfigure}
    \begin{subfigure}[c]{0.119\textwidth}
    \includegraphics[width=\linewidth]{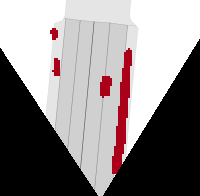}%
    \end{subfigure}
    \begin{subfigure}[c]{0.119\textwidth}
    \includegraphics[width=\linewidth]{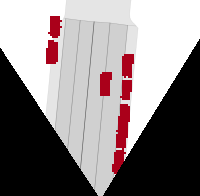}%
    \end{subfigure}
    \begin{subfigure}[c]{0.119\textwidth}
    \includegraphics[width=\linewidth]{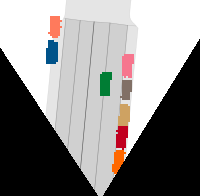}%
    \end{subfigure}

    \par\smallskip
    \begin{subfigure}[c]{0.25\textwidth}
    \centering
    \includegraphics[width=\linewidth]{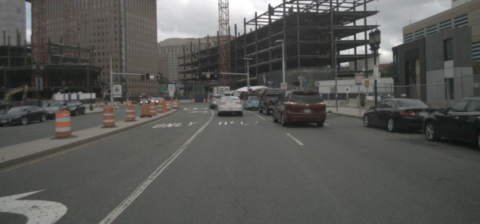}%
    \end{subfigure}
    \begin{subfigure}[c]{0.119\textwidth}
    \includegraphics[width=\linewidth]{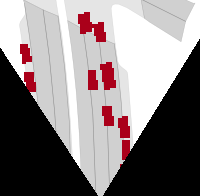}%
    \end{subfigure}
    \begin{subfigure}[c]{0.119\textwidth}
    \includegraphics[width=\linewidth]{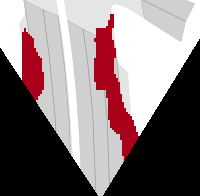}%
    \end{subfigure}
    \begin{subfigure}[c]{0.119\textwidth}
    \includegraphics[width=\linewidth]{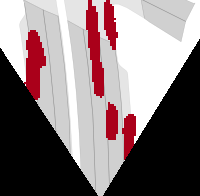}%
    \end{subfigure}
    \begin{subfigure}[c]{0.119\textwidth}
    \includegraphics[width=\linewidth]{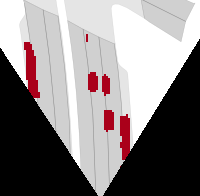}%
    \end{subfigure}
    \begin{subfigure}[c]{0.119\textwidth}
    \includegraphics[width=\linewidth]{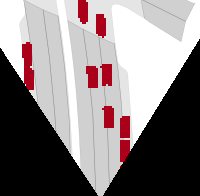}%
    \end{subfigure}
    \begin{subfigure}[c]{0.119\textwidth}
    \includegraphics[width=\linewidth]{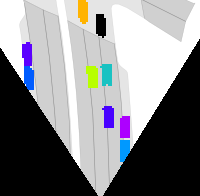}%
    \end{subfigure}

    \par\smallskip
    \begin{subfigure}[c]{0.25\textwidth}
    \centering
    \includegraphics[width=\linewidth]{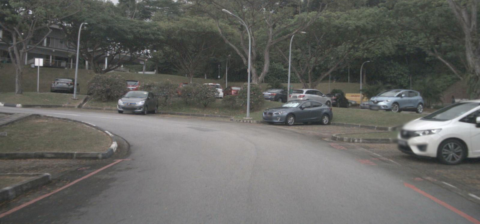}%
    \caption*{Camera input}
    \end{subfigure}
    \begin{subfigure}[c]{0.119\textwidth}
    \includegraphics[width=\linewidth]{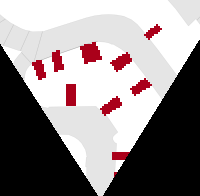}%
    \caption*{Ground truth}
    \end{subfigure}
    \begin{subfigure}[c]{0.119\textwidth}
    \includegraphics[width=\linewidth]{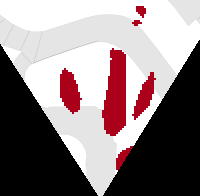}%
    \caption*{VPN \cite{Pan_2020}}
    \end{subfigure}
    \begin{subfigure}[c]{0.119\textwidth}
    \includegraphics[width=\linewidth]{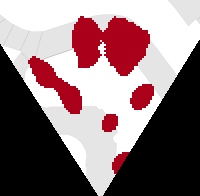}%
    \caption*{PON \cite{roddick20}}
    \end{subfigure}
    \begin{subfigure}[c]{0.119\textwidth}
    \includegraphics[width=\linewidth]{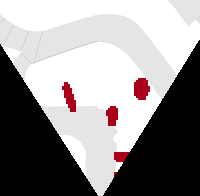}%
    \caption*{Lift-Splat \cite{philion20}}
    \end{subfigure}
    \begin{subfigure}[c]{0.119\textwidth}
    \includegraphics[width=\linewidth]{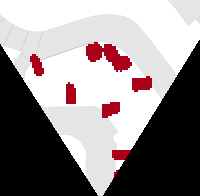}%
     \caption*{\textbf{Ours Seg.}}
    \end{subfigure}
    \begin{subfigure}[c]{0.119\textwidth}
    \includegraphics[width=\linewidth]{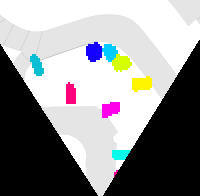}%
    \caption*{\textbf{Ours Inst.}}
    \end{subfigure}
\caption{Qualitative comparison of bird's-eye view prediction with published methods on NuScenes. The predictions of our model are much sharper and more accurate. Contrary to previous methods, FIERY can separate closely parked cars and correctly predict distant vehicles (near the top of the bird's-eye view image).}
\label{fig:qual-literature}
\vspace{-10pt}
\end{figure*}

\Cref{fig:qual-literature} shows a qualitative comparison of the predictions from our model with previous published methods, on the task of present-frame bird's-eye view semantic segmentation.

\subsection{Benefits of temporal fusion}
\begin{figure*}
\centering
    \begin{subfigure}[c]{0.95\textwidth}
    \centering
    \includegraphics[width=\linewidth]{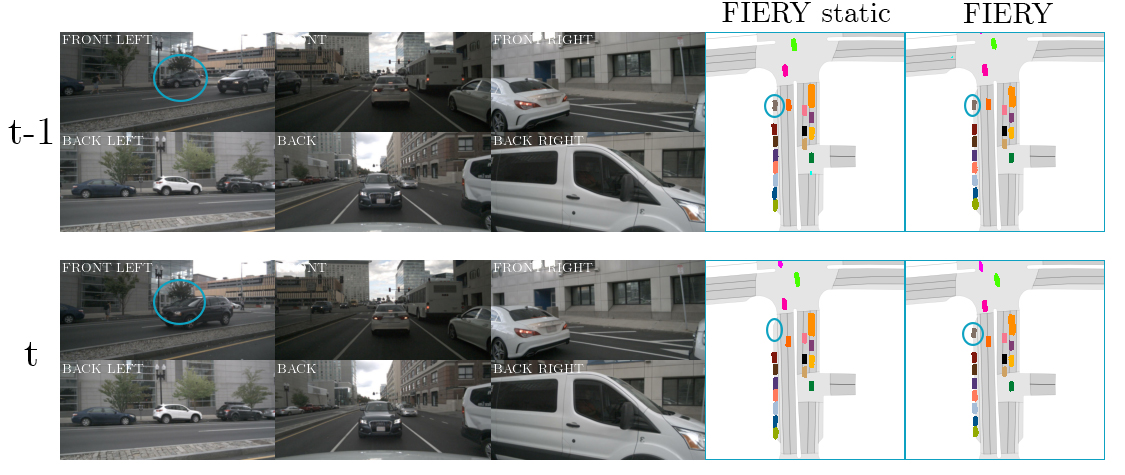}%
    \caption{The vehicle parked on the left-hand side is correctly predicted even through the occlusion.}
    \end{subfigure}
    \begin{subfigure}[c]{0.95\textwidth}
    \includegraphics[width=\linewidth]{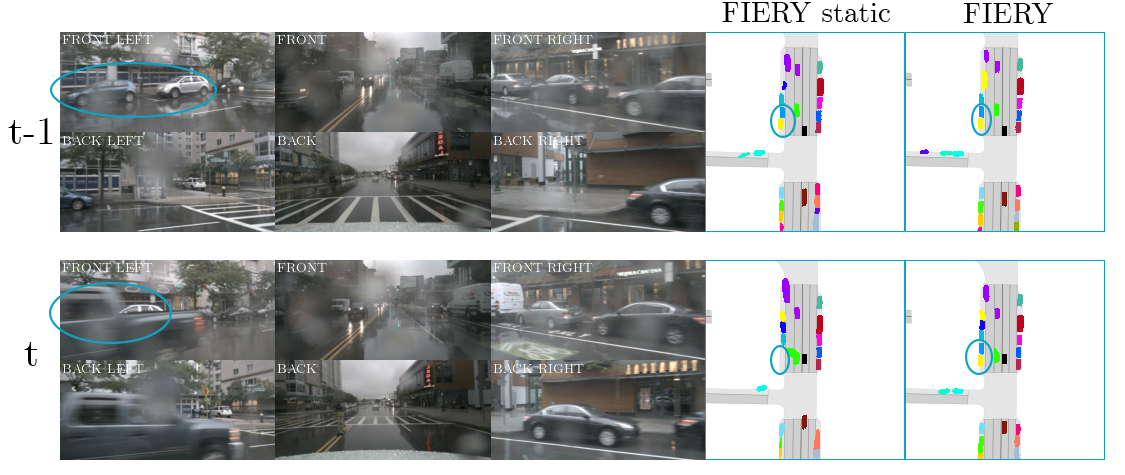}%
    \caption{The two vehicles parked on the left are heavily occluded by the 4x4 driving on the opposite lane, however by fusing past temporal information, the model is able to predict their positions accurately.}
    \end{subfigure}
\caption{Comparison of FIERY Static (no temporal context) and FIERY (1.0s of past context) on the task of present-frame bird's-eye view instance segmentation on NuScenes. FIERY can predict partially observable and occluded elements, as highlighted by the blue ellipses.}
\label{fig:temporal-fusion}
\vspace{-10pt}
\end{figure*}

When predicting the present-frame bird's-eye view segmentation, incorporating information from the past results in better predictions as shown in \Cref{fig:temporal-fusion}.

\subsection{Probabilistic modelling}
\label{appendix:probabilistic}
\paragraph{Generalised Energy Distance.}
Let $(\hat{Y}$, $\hat{Y}')$ be samples of predicted futures from our model, $(Y$, $Y')$ be samples of ground truth futures and $d$ be a distance metric. The Generalised Energy Distance $D_{\text{GED}}$ is defined as:

\begin{equation}
  D_{\text{GED}} = 2\mathbb{E} [d(\hat{Y},Y)] - \mathbb{E} [ d(\hat{Y},\hat{Y}')]  - \mathbb{E} [ d(Y,Y')]
\end{equation} 

We set our distance metric $d$ to $d(x,y) = 1- \text{VPQ}(x,y)$. Since we only have access to a unique ground truth future Y, $D_{\text{GED}}$ simplifies to:

\begin{equation}
  D_{\text{GED}} = 2\mathbb{E} [d(\hat{Y},Y)] - \mathbb{E} [ d(\hat{Y},\hat{Y}')]
\end{equation} 

\paragraph{Baselines.}
We describe below the baselines we compare our model to in \Cref{table:diversity}.
\begin{itemize}[itemsep=0.5mm, parsep=0pt]
    \item \textbf{M-Head.} The M-head model inspired by \cite{rupprecht17} outputs $M$ different futures. During training, the best performing head backpropagates its loss with weight $(1-\epsilon)$ while the other heads are weighted by $\frac{\epsilon}{M-1}$. We set $\epsilon=0.05$.
    \item \textbf{Bayesian Dropout.} We insert a dropout layer after every 3D temporal convolution in the temporal model. We also insert a dropout layer in the first 3 layers of the decoder, similarly to \cite{kendall17_segnet}. We set the dropout parameter to $p=0.25$.
    \item \textbf{Classical VAE.} We use a Centered Unit Gaussian to constrain our probability distribution similarly to the technique used in \cite{babaeizadeh18}. Different latent codes are sampled from $\mathcal{N}(0, I_L)$ during inference. 
\end{itemize}

\begin{table}[th!]
\centering
\begin{tabularx}{0.46\textwidth}{l|YY}
\thickhline
    \multicolumn{1}{c|}{} 
& \multicolumn{2}{c}{\color{darkgray}{\textbf{Generalised Energy Distance} ($\downarrow$)}}\\ 
& Short & Long \\
\mediumhline
M-Head & 96.6& 122.3 \\
Bayesian Dropout & 92.5& 116.5\\
Classical VAE & 93.2& 109.6\\
\textbf{FIERY} & \textbf{90.5}&  \textbf{105.1}\\
\thickhline
\end{tabularx}
\caption{Generalised Energy Distance on NuScenes, for $2.0$s future prediction and $M=10$ samples, showing that our model is able to predict the most accurate and diverse futures.}
\label{table:diversity}
\vspace{-10pt}
\end{table}

\clearpage

\subsection{Visualisation of the learned states}
\begin{figure*}[!b]
\centering
    \begin{subfigure}[c]{0.8\textwidth}
    \centering
    \includegraphics[width=\linewidth]{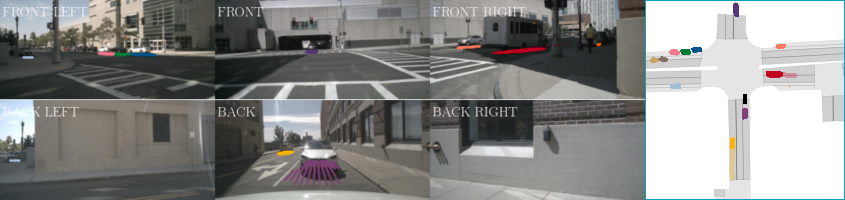}%
    \end{subfigure}
    \begin{subfigure}[c]{0.8\textwidth}
    \includegraphics[width=\linewidth]{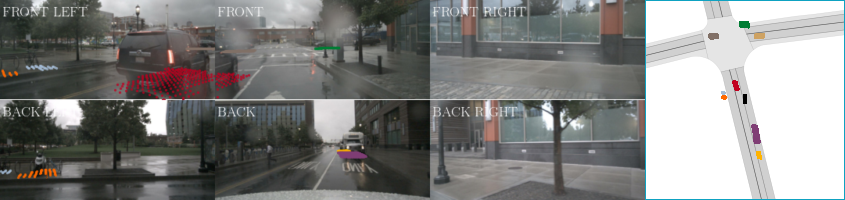}%
    \end{subfigure}
    \begin{subfigure}[c]{0.8\textwidth}
    \includegraphics[width=\linewidth]{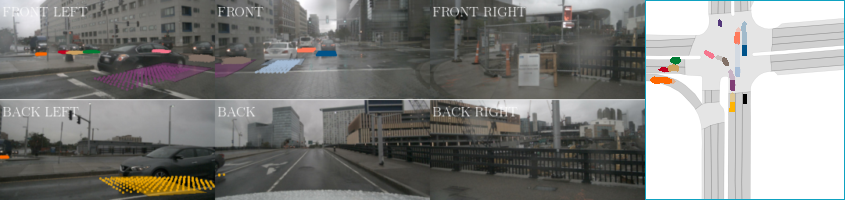}%
    \caption{Approaching an intersection.}
    \end{subfigure}
\caption{An example of cluster obtained from the spatio-temporal states $s_t$ by running a Gaussian Mixture algorithm on the NuScenes validation set. Our model learns to map similar situations to similar states. Even though the training dataset is deterministic, after mapping the RGB inputs to the state $s_t$, different futures can be observed from the same starting state. This explains why our probabilistic paradigm can learn to predict diverse and plausible futures.}
\label{fig:cluster1}
\vspace{-15pt}
\end{figure*}

\begin{figure*}
\centering
    \begin{subfigure}[c]{0.8\textwidth}
    \centering
    \includegraphics[width=\linewidth]{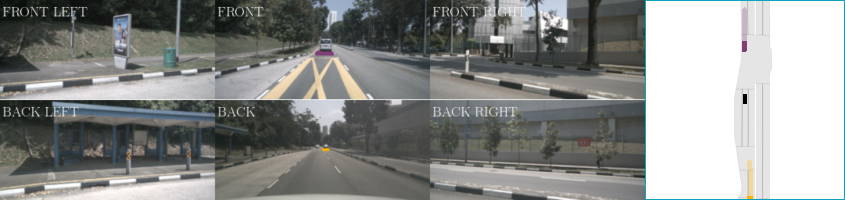}%
    \end{subfigure}
    \begin{subfigure}[c]{0.8\textwidth}
    \includegraphics[width=\linewidth]{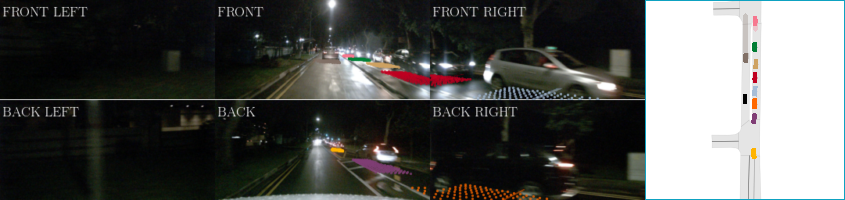}%
    \end{subfigure}
    \begin{subfigure}[c]{0.8\textwidth}
    \includegraphics[width=\linewidth]{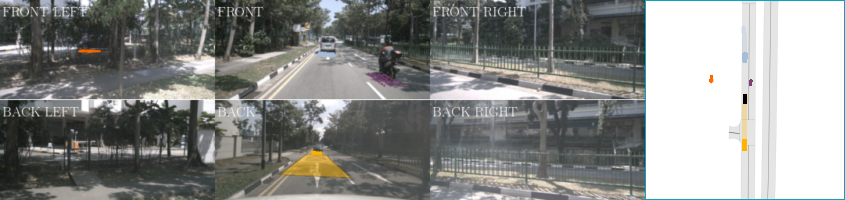}%
    \caption{Cruising behind a vehicle.}
    \end{subfigure}
    \par\smallskip
    \begin{subfigure}[c]{0.8\textwidth}
    \centering
    \includegraphics[width=\linewidth]{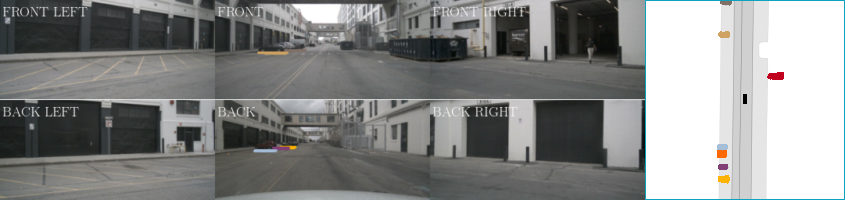}%
    \end{subfigure}
    \begin{subfigure}[c]{0.8\textwidth}
    \includegraphics[width=\linewidth]{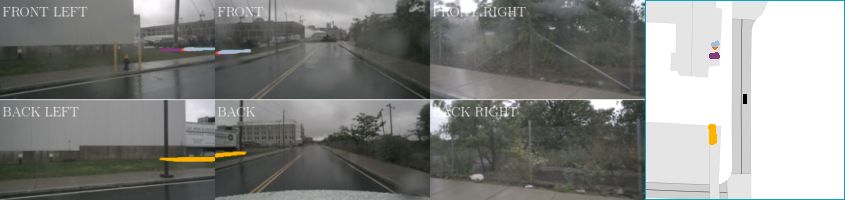}%
    \end{subfigure}
    \begin{subfigure}[c]{0.8\textwidth}
    \includegraphics[width=\linewidth]{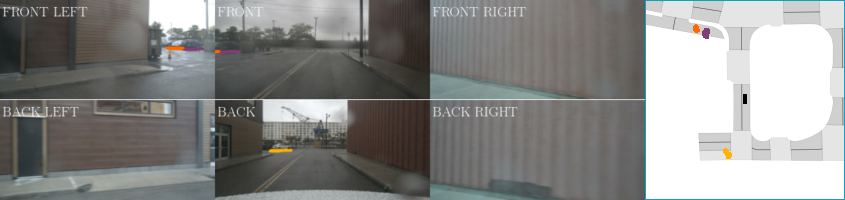}%
    \caption{Driving on open road.}
    \end{subfigure}
\caption{More example of clusters.}
\label{fig:cluster2-3}
\end{figure*}

\begin{figure*}
\centering
    \begin{subfigure}[c]{0.8\textwidth}
    \centering
    \includegraphics[width=\linewidth]{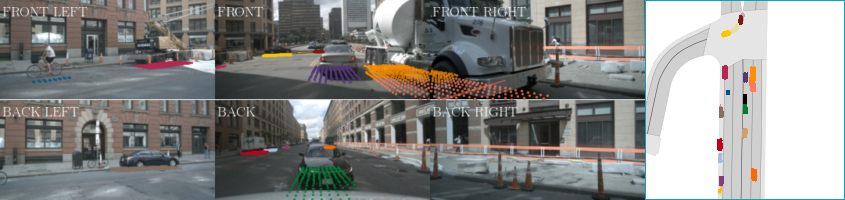}%
    \end{subfigure}
    \begin{subfigure}[c]{0.8\textwidth}
    \includegraphics[width=\linewidth]{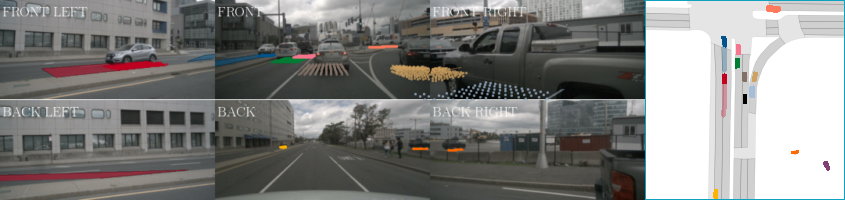}%
    \caption{Stuck in traffic.}
    \end{subfigure}
    \par\smallskip
    \begin{subfigure}[c]{0.8\textwidth}
    \centering
    \includegraphics[width=\linewidth]{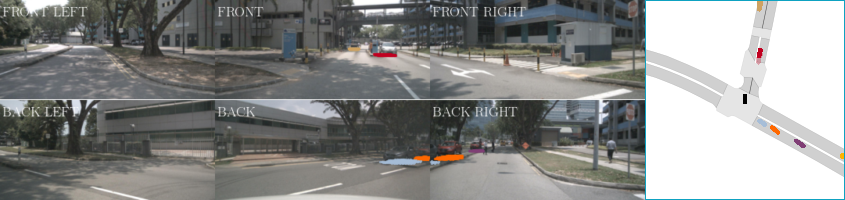}%
    \end{subfigure}
    \begin{subfigure}[c]{0.8\textwidth}
    \includegraphics[width=\linewidth]{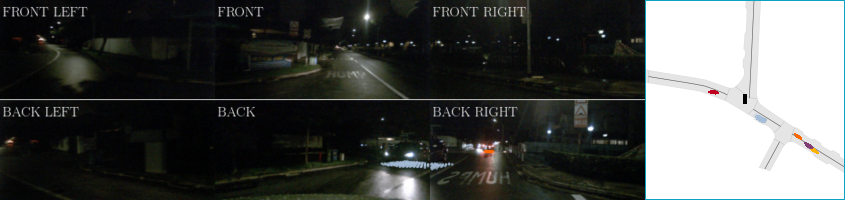}%
    \end{subfigure}
    \begin{subfigure}[c]{0.8\textwidth}
    \includegraphics[width=\linewidth]{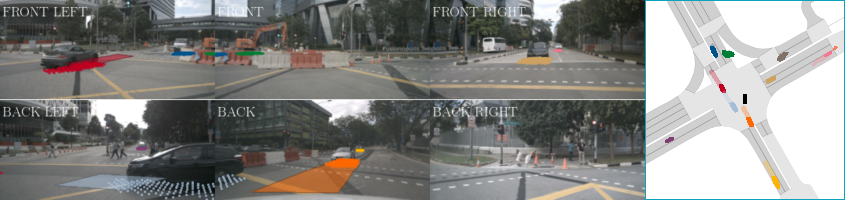}%
    \caption{Turning right at an intersection.}
    \end{subfigure}
\caption{More example of clusters.}
\label{fig:cluster4-5}
\end{figure*}

We run a Principal Component Analysis on the states $s_t$ and a Gaussian Mixture algorithm on the projected features in order to obtain clusters. We then visualise the inputs and predictions of the clusters in \Cref{fig:cluster1,fig:cluster2-3,fig:cluster4-5}. We observe that examples in a given cluster correspond to similar scenarios. Therefore, we better understand why our model is able to learn diverse and multimodal futures from a deterministic training dataset. Since similar scenes are mapped to the same state $s_t$, our model will effectively observe different futures starting from the same initial state. The present distribution will thus learn to capture the different modes in the future.

\subsection{Temporal horizon of future prediction}
\label{appendix:temporal-horizon}

\Cref{figure:temporal-horizon} shows the performance of our model for different temporal horizon: from 1.0s to 8.0s in the future. The performance seems to plateau beyond 6.0s in the future. In such a large future horizon, the prediction task becomes increasingly difficult as (i) uncertainty in the future grows further in time, and (ii) dynamic agents might not even be visible from past frames.

\definecolor{blue1}{RGB}{100,199,205}
\definecolor{blue2}{RGB}{30,154,214}
\definecolor{blue3}{RGB}{15,89,154}

\begin{figure}[h]
\centering
    \begin{tikzpicture}
    \begin{axis}[
        xlabel={Future time (s)},
        ylabel={Video Panoptic Quality},
        xmin=1.0, xmax=8.0,
        ymin=15, ymax=65,
        xtick={1,2,3,4,5,6, 7, 8},
        ytick={20, 30, 40, 50, 60},
        legend pos=north east,
        legend cell align={left},
        ymajorgrids=true,
        grid style=dashed,
    ]
        
    \addplot[
        color=blue2,
        mark=*,
        ]
        coordinates {
        (1.0,54.1)(2.0,49.7)(3.0,43.5)(4.0,39.6)(6.0,34.1)(8.0,34.1)
        };
    \addplot[
        color=blue3,
        mark=*,
        ]
        coordinates {
        (1.0,32.0)(2.0,29.5)(3.0,25.5)(4.0,23.6)(6.0,19.3)(8.0,18.9)
        };
    \addlegendentry{Short}
    \addlegendentry{Long}
        
    \end{axis}
    \end{tikzpicture}
\caption{Future prediction performance for different temporal horizons. We report future Video Panoptic Quality on NuScenes at different capture sizes around the ego-car: $30\mathrm{m}\times30\mathrm{m}$ (Short) and $100\mathrm{m}\times100\mathrm{m}$ (Long).}
\label{figure:temporal-horizon}
\end{figure}
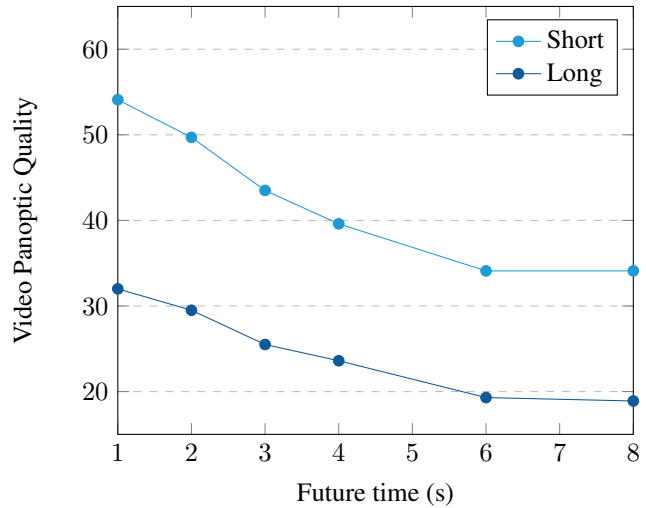

\clearpage
\clearpage
\section{Model and Dataset}
\label{appendix:model}

\subsection{Model description}
Our model processes $t=3$ past observations each with $n=6$ cameras images at resolution $(H_{\mathrm{in}}, W_{\mathrm{in}}) = (224\times480)$, i.e. $18$ images. 
The minimum depth value we consider is $D_{\text{min}} = 2.0\mathrm{m}$, which corresponds to the spatial extent of the ego-car. The maximum depth value is $D_{\text{max}}=50.0\mathrm{m}$, and the size of each depth slice is set to $D_{\mathrm{size}}=1.0\mathrm{m}$.

 We use uncertainty \cite{kendall18} to weight the segmentation, centerness, offset and flow losses. The probabilistic loss is weighted by $\lambda_{\text{probabilistic}} = 100$.

Our model contains a total of $8.1$M parameters and trains in a day on $4$ Tesla V100 GPUs with $32$GB of memory. All our layers use batch normalisation and a ReLU activation function.

\paragraph{Bird's-eye view encoder.} 
For every past timestep, each image in the observation $O_t = \{I_t^1, ..., I_t^n\}$ is encoded: $e_t^k = E(I_t^k) \in \mathbb{R}^{(C+D)\times H_e \times W_e}$. We use the EfficientNet-B4 \cite{tan19} backbone with an output stride of $8$ in our implementation, so $(H_e, W_e) = (\frac{H_{\mathrm{in}}}{8}, 
\frac{W_{\mathrm{in}}}{8}) = (28,60)$. The number of channel is $C=64$ and the number of depth slices is $D = \frac{D_{\mathrm{max}} - D_{\mathrm{min}}}{D_{\mathrm{size}}}=48$.

These features are then lifted and projected to bird's-eye view to obtain a tensor $x_t \in \mathbb{R}^{C\times H \times W}$ with $(H, W) = (200, 200)$. Using past ego-motion and a spatial transformer module, we transform the bird's-eye view features to the present's reference frame.

\paragraph{Temporal model.} The 3D convolutional temporal model is composed of \emph{Temporal Blocks}. Let $C_{\mathrm{in}}$ be the number of input channels and $C_{\mathrm{out}}$ the number of output channels. A single Temporal block is composed of:
\begin{itemize}[itemsep=0.3mm, parsep=0pt]
    \item a 3D convolution, with kernel size $(k_t, k_s, k_s) = (2, 3, 3)$. $k_t$ is the temporal kernel size, and $k_s$ the spatial kernel size.
    \item a 3D convolution with kernel size $(1, 3, 3)$.
    \item a 3D global average pooling layer with kernel size $(2, H, W)$.
\end{itemize}

Each of these operations are preceded by a feature compression layer, which is a $(1,1,1)$ 3D convolution with output channels $\frac{C_{\mathrm{in}}}{2}$.

All the resulting features are concatenated and fed through a $(1,1,1)$ 3D convolution with output channel $C_{\mathrm{out}}$. The temporal block module also has a skip connection. The final feature $s_t$ is in $\mathbb{R}^{64\times 200 \times 200}$. 

\paragraph{Present and future distributions.}
The architecture of the present and future distributions are identical, except for the number of input channels. The present distribution takes as input $s_t$, and the future distribution takes as input the concatenation of $(s_t, y_{t+1}, ..., y_{t+H})$. Let $C_p=64$ be the number of input channel of the present distribution and $C_f=64 + C_y\cdot H = 88$ the number of input channels of the future distribution (since $C_y=6$ and $H=4$). The module contains four residual block layers \cite{he16} each with spatial downsampling 2. These four layers divide the number of input channels by 2. A spatial average pooling layer then decimates the spatial dimension, and a final (1,1) 2D convolution regress the mean and log standard deviation of the distribution in $\mathbb{R}^L \times \mathbb{R}^L$ with $L=32$.

\paragraph{Future prediction.}
The future prediction module is made of the following structure repeated three times: a convolutional Gated Recurrent Unit \cite{ballas16} followed by $3$ residual blocks with kernel size $(3,3)$.

\paragraph{Future instance segmentation and motion decoder.}
The decoder has a shared backbone and multiple output heads to predict centerness, offset, segmentation and flow. The shared backbone contains:

\begin{itemize}[itemsep=0.3mm, parsep=0pt]
    \item a 2D convolution with output channel 64 and stride 2.
    \item the following block repeated three times: four 2D residual convolutions with kernel size (3, 3). The respective output channels are [64, 128, 256] and strides [1, 2, 2].
    \item three upsampling layers of factor 2, with skip connections and output channel 64.
\end{itemize}

Each head is then the succession two 2D convolutions outputting the required number of channels.

\subsection{Labels generation}
\label{appendix:dataset}
We compute instance center labels as a 2D Gaussian centered at each instance center of mass with standard deviation $\sigma_x = \sigma_y = 3$. The centerness label indicates the likelihood of a pixel to be the center of an instance and is a $\mathbb{R}^{1\times H\times W}$ tensor.
For all pixels belonging to a given instance, we calculate the offset labels as the vector pointing to the instance center of mass (a $\mathbb{R}^{2\times H\times W}$ tensor). Finally, we obtain future flow labels (a $\mathbb{R}^{2\times H\times W}$ tensor) by comparing the position of the instance centers of gravity between two consecutive timesteps.

We use the \emph{vehicles} category to obtain 3D bounding boxes of road agents, and filter the vehicles that are not visible from the cameras.

We report results on the official NuScenes validation split. Since the Lyft dataset does not provide a validation set, we create one by selecting random scenes from the dataset so that it contains roughly the same number of samples (6,174) as NuScenes (6,019).

